\documentclass[11pt,a4paper]{article}
\usepackage[hyperref]{emnlp2018}
\usepackage{hyperref}
\usepackage{times}
\usepackage{latexsym}
\usepackage{graphicx} 
\usepackage{mathtools}
\usepackage{subfigure}
\usepackage{helvet}
\usepackage{courier}
\usepackage{bm}
\usepackage{booktabs,multirow}
\usepackage{bigstrut,bigdelim}
\usepackage{amsfonts}
\usepackage{amsmath}
\usepackage{mathrsfs}
\usepackage{float}
\usepackage{multirow}
\usepackage{xcolor,colortbl} 
\usepackage{url}

\aclfinalcopy 

\title{Iterative Document Representation Learning Towards Summarization with Polishing}

\author{Xiuying Chen$^1$, Shen Gao$^2$, Chongyang Tao$^2$, Yan Song$^{3}$, Dongyan Zhao$^{1,2}$ and Rui Yan$^{1,2,}$\thanks{ Corresponding author: Rui Yan (ruiyan@pku.edu.cn) }\\
 $^1$Center for Data Science, Peking University, Beijing, China\\
 $^2$Institute of Computer Science and Technology, Peking University, Beijing, China\\
 $^3$Tencent AI Lab\\
 \tt \{xy-chen,shengao,chongyangtao,zhaody,ruiyan\}@pku.edu.cn\\
 \tt clksong@tencent.com
  }

\date{}

\begin{document}
\maketitle
\begin{abstract}
In this paper, we introduce Iterative Text Summarization (ITS), an iteration-based model for supervised extractive text summarization, inspired by the observation that it is often necessary for a human to read an article multiple times in order to fully understand and summarize its contents. Current summarization approaches read through a document only once to generate a document representation, resulting in a sub-optimal representation. To address this issue we introduce a model which iteratively polishes the document representation on many passes through the document. As part of our model, we also introduce a selective reading mechanism that decides more accurately the extent to which each sentence in the model should be updated. Experimental results on the CNN/DailyMail and DUC2002 datasets demonstrate that our model significantly outperforms state-of-the-art extractive systems when evaluated by machines and by humans.
\end{abstract}

\section{Introduction}

A summary is a shortened version of a text document which maintains the most important ideas from the original article. Automatic text summarization is a process by which a machine gleans the most important concepts from an article, removing secondary or redundant concepts. 
Nowadays as there is a growing need for storing and digesting large amounts of textual data, automatic summarization systems have significant usage potential in society.

\textit{Extractive summarization} is a technique for generating summaries by directly choosing a subset of salient sentences from the original document to constitute the summary. 
Most efforts made towards extractive summarization either rely on human-engineered features such as sentence length, word position, and frequency \cite{Cohen2002Natural,radev2004mead,Woodsend2010Automatic,Yan2011Summarize,Yan2011Evolutionary,yan2012visualizing} or use neural networks to automatically learn features for sentence selection \cite{Cheng2016Neural,Nallapati2016SummaRuNNer}. 

Although existing extractive summarization methods have achieved great success, one limitation they share is that they generate the summary after only one pass through the document.
However, in real-world human cognitive processes, people read a document multiple times in order to capture the main ideas. 
Browsing through the document only once often means the model cannot fully get at the document's main ideas, leading to a subpar summarization. We share two examples of this. (1) Consider the situation where we almost finish reading a long article and forget some main points in the beginning. We are likely to go back and review the part that we forget.
(2) To write a good summary, we usually first browse through the document to obtain a general understanding of the article, then perform a more intensive reading to select salient points to include in the summary. 
In terms of model design, we believe that letting a model read through a document multiple times, polishing and updating its internal representation of the document can lead to better understanding and better summarization.

	
To achieve this, we design a model that we call Iterative Text Summarization (ITS) consisting of a novel ``iteration mechanism'' and ``selective reading module''.
ITS is an iterative process, reading through the document many times.
There is one encoder, one decoder, and one iterative unit in each iteration. They work together to polish document representation. The final labeling part uses outputs from all iterations to generate summaries.
The selective reading module we design is a modified version of a Gated Recurrent Unit (GRU) network, which can decide how much of the hidden state of each sentence should be retained or updated based on its relationship with the document.

Overall, our contribution includes: 
\begin{enumerate}
	\item  We propose Iterative Text Summarization (ITS), an iteration based summary generator which uses a sequence classifier to extract salient sentences from documents.
	\item We introduce a novel iterative neural network model which repeatedly polishes the distributed representation of document instead of generating that once for all. 
	Besides, we propose a selective reading mechanism, which decides how much information should be updated of each sentence based on its relationship with the polished document representation.
	Our entire architecture can be trained in an end-to-end fashion. 
	\item We evaluate our summarization model on representative CNN/DailyMail corpora and benchmark DUC2002 dataset. Experimental results demonstrate that our model outperforms state-of-the-art extractive systems when evaluated automatically and by human.
	
\end{enumerate}


\section{Related Work}

Our research builds on previous works in two fields: summarization and iterative modeling.

Text summarization can be classified into extractive summarization and abstractive summarization. Extractive summarization aims to generate a summary by integrating the most salient sentences in the document. Abstractive summarization aims to generate new content that concisely paraphrases the document from scratch.

With the emergence of powerful neural network models for text processing, a vast majority of the literature on document summarization is dedicated to abstractive summarization. These models typically take the form of convolutional neural networks (CNN) or recurrent neural networks (RNN).
For example, \citet{Rush2015A} propose an encoder-decoder model which uses a local attention mechanism to generate summaries.
\citet{Nallapati2016Abstractive} further develop this work by addressing problems that had not been adequately solved by the basic architecture, such as keyword modeling and capturing the hierarchy of sentence-to-word structures.
In a follow-up work, \citet{Nallapati2017SenGen} propose a new summarization model which generates summaries by sampling a topic one sentence at a time, then producing words using an RNN decoder conditioned on the sentence topic. 
Another related work is by \citet{DBLP:journals/corr/SeeLM17}, where the authors use ``pointing'' and ``coverage'' techniques to generate more accurate summaries.

Despite the focus on abstractive summarization, extractive summarization remains an attractive method as it is capable of generating more grammatically and semantically correct summaries. This is the method we follow in this work. 
In extractive summarization,
\citet{Cheng2016Neural} propose a general framework for single-document text summarization using a hierarchical article encoder composed with an attention-based extractor. 
Following this, \citet{Nallapati2016SummaRuNNer} propose a simple RNN-based sequence classifier which outperforms or matches the state-of-art models at the time. 
In another approach, \citet{Narayan2018Ranking} use a reinforcement learning method to optimize the Rouge evaluation metric for text summarization.
The most recent work on this topic is \cite{wu2018learning}, where the authors train a reinforced neural extractive summarization model called RNES that captures cross-sentence coherence patterns. Due to the fact that they use a different dataset and have not released their code, we are unable to compare our models with theirs.


The idea of iteration has not been well explored for summarization. 
One related study is \citet{Xiong2016Dynamic}'s work on dynamic memory networks, which designs neural networks with memory and attention mechanisms that exhibit certain reasoning capabilities required for question answering. 
Another related work is \cite{yan2016poet}, where they generate poetry with iterative polishing sn
chema. Similiar method can also be applied on couplet generation as in \cite{Yan2016Chinese}.
We take some inspiration from their work but focus on document summarization.
Another related work is \cite{Singh2017Hybrid}, where the authors present a deep network called Hybrid MemNet for the single document summarization task, using a memory network as the document encoder. 
Compared to them, we do not borrow the memory network structure but propose a new iterative architecture.


\section{Methodology}

\subsection{Problem Formulation}
In this work, we propose Iterative Text Summarization (ITS), an iteration-based supervised model for extractive text summarization. 
We treat the extractive summarization task as a sequence labeling problem, in which each sentence is visited sequentially and a binary label that determines whether or not it will be included in the final summary is generated.

ITS takes as input a list of sentences $ s = \{s_{1},\dots,s_{n_s}\}$, where $n_{s}$ is the number of sentences in the document. Each sentence $s_{i}$ is a list of words: $s_{i} = \{w_1^i,\dots,w_{n_w}^i\}$, where $n_w$ is the word length of the sentence.
The goal of ITS is to generate a score vector $\bm{y} = \{y_{1},\dots,y_{n_s}\}$ for each sentence, where each score $y_{i} \in [0,1]$ denotes the sentence's \textit{extracting probability}, that is, the probability that the corresponding sentence $s_i$ will be extracted to be included in the summary. 
We train our model in a supervised manner, using a corresponding gold summary written by human experts for each document in training set. We use an unsupervised method to convert the human-written summaries to gold label vector $\bm{y'}=\{y'_{1},...,y'_{n_{s}}\}$, where $y'_{i}\in \{0,1\}$ denotes whether the $i$-th sentence is selected (1) or not (0).
Next, during training process, the cross entropy loss is calculated between $\bm{y}$ and $\bm{y'}$, which is minimized to optimize $\bm{y}$.
Finally, we select three sentences with the highest score according to $\bm{y}$ to be the extracted summary. 
We detail our model below.

\subsection{Model Architecture}

\begin{figure*}
	\setlength{\abovecaptionskip}{0.cm}
	
	\setlength{\belowcaptionskip}{-0.cm}
	\includegraphics[width=2\columnwidth]{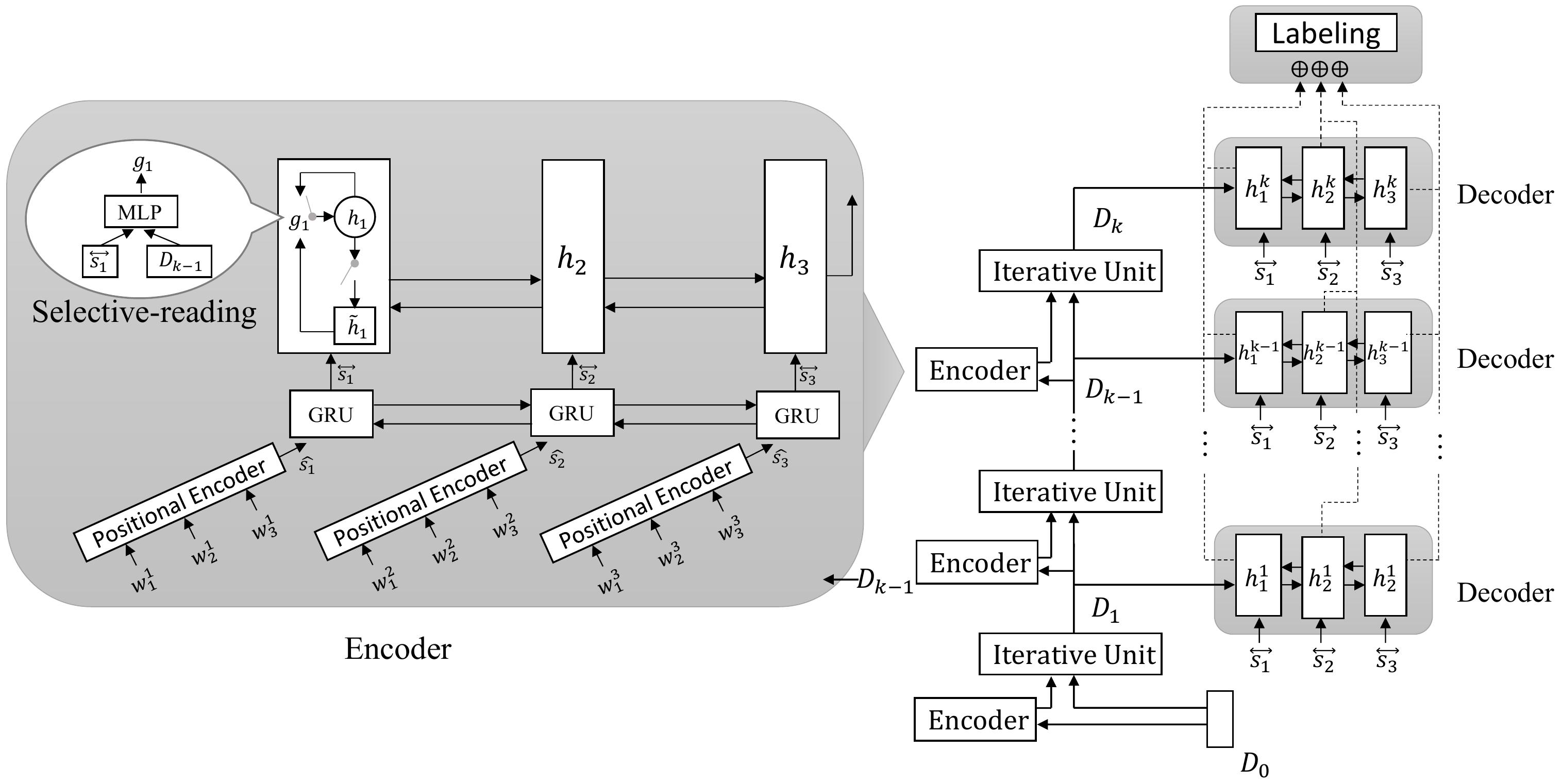}
	\caption{\label{fig:overall} Model Structure: There is one encoder, one decoder and one iterative unit  (which is used to polish document representation) in each iteration. The final labeling part is used to generating the extracting probabilities for all sentences combining hidden states of decoders in all iterations. We take a document consists of three sentences for example here. }
\end{figure*}

ITS is depicted in Fig.\ref{fig:overall}. It consists of multiple iterations with one encoder, one decoder, and one iteration unit in each iteration. We combine the outputs of decoders in all iterations to generate the extracting probabilities in the final labeling module.

Our encoder is illustrated in the shaded region in the left half of Fig.\ref{fig:overall}. It takes as input all sentences as well as the document representation from the previous unit $\bm{D_{k-1}}$, processes them through several neural networks, and outputs the final state to the iterative unit module which updates the document representation. 

Our decoder takes the form of a bidirectional RNN. It takes the representation of sentence generated by the encoder as input, and its initial state is the polished document representation $\bm{D_{k}}$. 
Our last module, the sentence labeling module, concatenates the hidden states of all decoders together to generate an integrated score for each sentence.

As we apply supervised training, the objective is to maximize the likelihood of all sentence labels $y'=\{y_{1}',...,y_{n_{s}}'\}$ given the input document $s$ and model parameters $\theta$:

\begin{equation}
\log p(y'|s;\theta)=\sum_{i=1}^{n_s}{\log p(y_{i}'|s;\theta)}
\end{equation}

\section{Our Model}

\subsection{Encoder}

In this subsection, we describe the encoding process of our model.
For brevity, we drop the superscript k when focusing on a particular layer.
All the $\bm{W}$'s and $\bm{b}$'s in this section  with different superscripts or subscripts are the parameters to be learned.

\textbf{Sentence Encoder: } 
Given a discrete set of sentences $s = \{s_{1},\dots,s_{n_s}\}$, we use a word embedding matrix $\bm{M} \in \mathbb{R}^{V \times D}$ to embed each word $w^{i}$ in sentence $s_{i}$ into continuous space $\bm{\hat{w}^{i}}$, where $V$ is the vocabulary size, $D$ is the dimension of word embedding.

The sentence encoder can be based on a variety of encoding schemes.
Simply taking the average of embeddings of words in a sentence will cause too much information loss, while using GRUs or Long Short-Term Memory (LSTM) requires more computational resources and is prone to overfitting. 
Considering above, we select \textit{positional encoding} described in \cite{sukhbaatar2015end} as our sentence encoding method.
Each sentence representation $\bm{\hat{s_{i}}}$ is calculated by $\bm{\hat{s_{i}}}=\sum_{j=1}^{n_{w}}\bm{l_{j}} \circ \bm{\hat{w}^{i}_{j}}$, where $\circ$ is element-wise multiplication, 
$\bm{l_{j}}$ is a column vector computed as $l_{j,d}=(1-\frac{j}{n_{w}})-(\frac{d}{D})(1-\frac{2j}{n_{w}})$, $l_{j,d}$ denotes the $d$-th dimension of $\bm{l_{j}}$. 

Note that throughout this study, we use GRUs as our RNN cells since they can alleviate the overfitting problem as confirmed by our experiments.
As our selective reading mechanism (which will be explained later) is a modified version of original GRU cell, we give the details of the GRU here.
GRU is a gating mechanism in recurrent neural networks, introduced in \cite{Cho2014Learning}. Their performance was found to be similar to that of LSTM cell but using fewer parameters as described in \cite{hochreiter1997long}.
The GRU cell consists of an update gate vector $\bm{u_i}$, a reset gate vector $\bm{r_i}$, and an output vector $\bm{h_i}$.
For each time step $i$ with input $\bm{x_{i}}$ and previous hidden state $\bm{h_{i-1}}$, the updated hidden state $\bm{h_{i}}=\text{GRU}(\bm{x_{i}},\bm{h_{i-1}})$ is computed by:
\begin{align}
\bm{u_{i}} &=\sigma(\bm{W^{(u)}}\bm{x_{i}}+\bm{U^{(u)}}\bm{h_{i-1}}+\bm{b^{(u)}})  \label{gated} \\
\bm{r_{i}} &=\sigma(\bm{W^{(r)}}\bm{x_{i}}+\bm{U^{(r)}}\bm{h_{i-1}}+\bm{b^{(r)}})  \\
\bm{\tilde{h_{i}}}&=\text{tanh}(\bm{W^{(h)}x_{i}}+\bm{r_{i}}\circ \bm{Uh_{i-1}}+\bm{b^{(h)}}) \\
\bm{h_{i}}&=\bm{u_{i}}\circ \bm{\tilde{h_{i}}}+(1-\bm{u_{i}})\circ \bm{h_{i-1} } \label{hi}
\end{align}
where $\sigma$ is the sigmoid activation function, $\bm{W^{(u)}}, \bm{W^{(r)}}, \bm{W^{(h)}} \in \mathbb{R}^{n_{H}\times n_{I}},\bm{U^{(u)}},\bm{U^{(r)}},\bm{U}\in \mathbb{R}^{n_{H}\times n_{H}}, n_{H}$ is the hidden size, $n_{I}$ is the size of input $\bm{x_{i}}$.

To further study the interactions and information exchanges between sentences, we establish a Bi-directional GRU (Bi-GRU) network taking the sentence representation as input:
\begin{align}
\bm{\overrightarrow{s_{i}}} & =\text{GRU}_{\text{fwd}}(\bm{\hat{s}_{i}},\bm{\overrightarrow{s_{i-1}}}) \\
\bm{\overleftarrow{s_{i}}} & =\text{GRU}_{\text{bwd}}(\bm{\hat{s}_{i}},\bm{\overleftarrow{s_{i-1}}}) \\
\bm{\overleftrightarrow{s_{i}}}& =\bm{\overrightarrow{s_{i}}}+\bm{\overleftarrow{s_{i}}}
\end{align}
where $\bm{\hat{s}_{i}}$ is the sentence representation input at time step $i$, $\overrightarrow{\bm{s_{i}}}$ is the hidden state of the forward GRU at time step $i$, and $\overleftarrow{\bm{s_{i}}}$ is the hidden state of the backward GRU.
This architecture allows information to flow back and forth to generate new sentence representation $\bm{\overleftrightarrow{s_{i}}}$.

\textbf{Document Encoder: } 
We must initialize a document representation before polishing it.
Generating the document representation from sentence representations is a process similar to generating the sentence representation from word embeddings. This time we need to compress the whole document, not just a sentence, into a vector.  
Because the information a vector can contain is limited, rather than to use another neural network, we simply use a non-linear transformation of the average pooling of the concatenated hidden states of the above Bi-GRU to generate the document representation, as written below:
\begin{align}
\bm{D_{0}}=\text{tanh}(\bm{W}\frac{1}{n_{s}}\sum^{n_{s}}_{i=1}[\bm{\overrightarrow{s_{i}}};\bm{\overleftarrow{s_{i}}}]+\bm{b})
\end{align}
where `[$\cdot$;$\cdot$]' is the concatenation operation.

\textbf{Selective Reading module: } 
Now we can formally introduce the selective reading module in Fig.\ref{fig:overall}. 
This module is a bidirectional RNN consisting of modified GRU cells whose input is the sentence representation $\bm{\overleftrightarrow{s}}=\{\bm{\overleftrightarrow{s_{1}}},...,\bm{\overleftrightarrow{s_{n_{s}}}}\}$.
In the original version of GRU, the update gate $\bm{u_{i}}$ in Equation \ref{gated} is used to decide how much of hidden state should be retained and how much should be updated. 
However, due to the way $\bm{u_{i}}$ is calculated, it is sensitive to the position and ordering of sentences, but loses information captured by the polished document representation.

Herein, we propose a modified GRU cell that replace the $\bm{u_{i}}$ with the newly computed update gate $\bm{g_{i}}$. The new cell takes in two inputs, the sentence representation and the document representation from the last iteration, rather than merely the sentence representation. For each sentence, the selective network generates an update gate vector $\bm{g_{i}}$ in the following way:
\begin{align}
\bm{f_{i}} & =[\bm{\overleftrightarrow{s_{i}}}\circ \bm{D_{k-1}}; \bm{\overleftrightarrow{s_{i}}}; \bm{D_{k-1}}] \\
\bm{F_{i}}& =\bm{W^{(2)}}\text{tanh}(\bm{W^{(1)}}\bm{f_{i}}+\bm{b^{(1)}})+\bm{b^{(2)}} \\
\bm{g_{i}}& =\frac{\text{exp}(\bm{F_{i}})}{\sum^{n_{s}}_{j=1}\text{exp}(\bm{F_{j}})}
\end{align}
where  $\bm{\overleftrightarrow{s_{i}}}$ is the $i$-th sentence representation, $\bm{D_{k-1}}$ is the document representation from last iteration. Equation \ref{hi} now becomes:
\begin{align}
\bm{h_{i}}&=\bm{g_{i}}\circ \bm{\tilde{{h_{i}}}}+(1-\bm{g_{i}})\circ \bm{h_{i-1} }
\end{align}

We use this ``selective reading module'' to automatically decide to which extent the information of each sentence should be updated based on its relationship with the polished document. In this way, the modified GRU network can grasp more accurate information from the document.

\subsection{Iterative Unit}
After each sentence passes through the selective reading module, we wish to update the document representation $\bm{D_{k-1}}$ with the newly constructed sentence representations. 
The iterative unit (also depicted above in Fig.\ref{fig:overall}) is designed for this purpose.
We use a $\text{GRU}_{\text{iter}}$ cell to generate the polished document representation, whose input is the final state of the selective reading network from the previous iteration, $\bm{h_{n_{s}}}$ and whose initial state is set to the document representation of the previous iteration, $\bm{D_{k-1}}$.  The updated document representation is computed by:
\begin{align}
\bm{D_{k}}&=\text{GRU}_{\text{iter}}(\bm{h_{n_{s}}},\bm{D_{k-1}})
\end{align}

\subsection{Decoder}
Next, we describe our decoders, which are depicted shaded in the right part of Fig.\ref{fig:overall}. 
Following most sequence labeling task \cite{xue2004calibrating,Carreras:2005:ICS:1706543.1706571} where they learn a feature vector for each sentence, we use a bidirectional $\text{GRU}_{\text{dec}}$ network in each iteration to output features so as to calculate extracting probabilities.
For $k$-th iteration, given the sentence representation $\bm{\overleftrightarrow{s}}$ as input and the document representation $\bm{D_k}$ as the initial state,
our decoder encodes the features of all sentences in the hidden state $\bm{h^{k}}=\{\bm{h^{k}_{0}},...,\bm{h^{k}_{n_{s}}}\}$:
\begin{align}
\bm{h^{k}_{i}}&= \text{GRU}_{\text{dec}}(\bm{\overleftrightarrow{s}}, \bm{h^{k}_{i-1}})\\
\bm{h^{k}_{0}}&=\bm{D_k}
\end{align}

\subsection{Sentence Labeling Module}
Next, we use the feature of each sentence to generate corresponding extracting probability. Since we have one decoder in each iteration, if we directly transform the hidden states in each iteration to extracting probabilities, we will end up with several scores for each sentence.
Either taking the average or summing them together by specific weights is inappropriate and inelegant.
Hence, we concatenate hidden states of all decoders together and apply a multi-layer perceptron to them to generate the extracting probabilities: 
\begin{equation}
\bm{y}=\bm{W^{(4)}}\text{tanh}(\bm{W^{(3)}}[\bm{h^{1}};...;\bm{h^{k}}]+\bm{b^{(3)}})+\bm{b^{(4)}}
\end{equation}
where $\bm{y}=\{y_{1},...,y_{n_{s}}\}$, $y_{i}$ is the extracting probability for each setence.
In this way, we let the model learn by itself how to utilize the outputs of all iterations and assign to each hidden state a reliable weight.
In section~\ref{analysis}, we will show that this labeling method outperforms other methods.

\section{Experiment Setup}
In this section, we present our experimental setup for training and estimating our summarization model.
We first introduce the datasets used for training and evaluation, and then introduce our experimental details and evaluation protocol.
\subsection{Datasets}
In order to make a fair comparison with our baselines, we used the CNN/Dailymail corpus which was constructed by \citet{Hermann2015Teaching}.
We used the standard splits for training, validation and testing in each corpus (90,266/1,220/1,093 documents for CNN and 196,557/12,147/10,396 for DailyMail).
We followed previous studies in using the human-written story highlight in each article as a gold-standard abstractive summary.
These highlights were used to generate gold labels when training and testing our model using the greedy search method similar to  \cite{Nallapati2016SummaRuNNer}.

We also tested ITS on an out-of-domain corpus, DUC2002, which consists of 567 documents.
Documents in this corpus belong to 59 various clusters and each cluster has a unique topic. 
Each document has two gold summaries written by human experts of length around 100 words.

\subsection{Implementation Details}
\def\UrlBreaks{\do\A\do\B\do\C\do\D\do\E\do\F\do\G\do\H\do\I\do\J\do\K\do\L\do\M\do\N\do\O\do\P\do\Q\do\R\do\S\do\T\do\U\do\V\do\W\do\X\do\Y\do\Z\do\[\do\\\do\]\do\^\do\_\do\`\do\a\do\b\do\c\do\d\do\e\do\f\do\g\do\h\do\i\do\j\do\k\do\l\do\m\do\n\do\o\do\p\do\q\do\r\do\s\do\t\do\u\do\v\do\w\do\x\do\y\do\z\do\0\do\1\do\2\do\3\do\4\do\5\do\6\do\7\do\8\do\9\do\.\do\@\do\\\do\/\do\!\do\_\do\|\do\;\do\>\do\]\do\)\do\,\do\?\do\'\do+\do\=\do\#}

We implemented our model in Tensorflow \cite{abadi2016tensorflow}. The code for our models is available online\footnote{\url{https://github.com/yingtaomj/Iterative-Document-Representation-Learning-Towards-Summarization-with-Polishing}}.
We mostly followed the settings in \cite{Nallapati2016SummaRuNNer} and trained the model using the Adam optimizer~\cite{Kingma2014Adam} with initial learning rate 0.001 and anneals of 0.5 every 6 epochs until reaching 30 epochs.
We selected three sentences with highest scores as summary. After preliminary exploration, we found that arranging them according to their scores consistently achieved the best performance.
Experiments were performed with a batch size of 64 documents. 
We used 100-dimension GloVe \cite{Pennington2014Glove} embeddings trained on Wikipedia 2014 as our embedding initialization with a vocabulary size limited to 100k for speed purposes.
We initialized out-of-vocabulary word embeddings over a uniform distribution within [-0.2,0,2].
We also padded or cut sentences to contain exactly 70 words. 
Each GRU module had 1 layer with 200-dimensional hidden states and with either an initial state set up as described above or a random initial state. 
To prevent overfitting, we used dropout after each GRU network and embedding layer, and also applied L2 loss to all unbiased variables.  
The iteration number was set to 5 if not specified. A detailed discussion about iteration number can be found in section~\ref{greetings}.

\subsection{Baselines}

On all datasets we used the Lead-3 method as a baseline, which simply chooses the first three sentences in a document as the gold summary.
On DailyMail datasets, we report the performance of SummaRuNNer in \cite{Nallapati2016SummaRuNNer} and the model in  \cite{Cheng2016Neural}, as well as a logistic regression classifier (LReg) that they used as a baseline.
We reimplemented the Hybrid MemNet model in \cite{Singh2017Hybrid} as one of our baselines since they only reported the performance of 500 samples in their paper.
Also, \citet{Narayan2018Ranking} released their code\footnote{\url{https://github.com/EdinburghNLP/Refresh}} for the REFRESH model, we used their code to produce Rouge recall scores on the DailyMail dataset as they only reported results on CNN/DailyMail joint dataset.
Baselines on CNN dataset are similar. 
On DUC2002 corpus, we compare our model with several baselines such as Integer Linear Programming (ILR) and LReg. 
We also report the performance of the newest neural networks model including \cite{Nallapati2016SummaRuNNer,Cheng2016Neural,Singh2017Hybrid}.

\subsection{Evaluation}
In the evaluation procedure, we used the Rouge scores, i.e. Rouge-1, Rouge-2, and Rouge-L, corresponding to the matches of unigram, bigrams, and Longest Common Subsequence (LCS) respectively, to estimate our model.
We obtained our Rouge scores using the standard pyrouge package\footnote{\url{https://pypi.python.org/pypi/pyrouge/0.1.0}}.
To compare with other related works, we used full-length F1 score on the CNN corpus, limited length of 75 bytes and 275 bytes recall score on DailyMail corpus. 
As for the DUC2002 corpus, following the official guidelines, we examined the Rouge recall score at the length of 75 words. 
All results in our experiment are statistically significant using 95\% confidence interval as estimated by Rouge script.

\citet{schluter2017limits} noted that only using the Rouge metric to evaluate summarization quality can be misleading. Therefore, we also evaluated our model using human evaluation. 
Five highly educated participants were asked to rank 40 summaries produced by four models: the Lead-3 baseline, Hybrid MemNet, ITS, and human-authored highlights. 
We chose Hybrid MemNet as one of the human evaluation baselines since its performance is relatively high compared to other baselines.
Judging criteria included informativeness and coherence. Test cases were randomly sampled from DailyMail test set.


\section{Experiment analysis}
\label{analysis}
\begin{table*}[t!]
	\setlength{\abovecaptionskip}{0.cm}
	
	\setlength{\belowcaptionskip}{-0.cm}
	\begin{center}
		\begin{tabular}{l|ccc|ccc}
			\hline \multirow{2}*{\large{DailyMail}}
			&& \small{b75} & & &  \small{b275} &\\
			\cline{2-7}
			 &  Rouge-1 &  Rouge-2 &  Rouge-L&  Rouge-1 &  Rouge-2 &  Rouge-L\\ 
			\hline
			Lead-3 & 21.9 & 7.2 & 11.6 & 40.5 & 14.9 & 32.6\\
			LReg(500) & 18.5 & 6.9 & 10.2& - & - & - \\
			Cheng \emph{et.al}'16  & 22.7 & 8.5 & 12.5 & 42.2 & 17.3 & \bf 34.8\\
			SummaRuNNer & 26.2 & 10.8 & 14.4& 42 & 16.9 & 34.1 \\
			REFRESH &24.1 & 11.5& 12.5& 40.3 & 15.1& 32.9\\
			Hybrid MemNet&26.3 & 11.2 & 15.5& 41.4 & 16.7 & 33.2\\
			ITS & \bf 27.4 & \bf 11.9 & \bf 16.1 & \bf42.4 & \bf17.4 & 34.1\\
			\hline
		\end{tabular}
	\end{center}
	\caption{\label{tab:compare-baseline} Comparison with other baselines on \textbf{DailyMail} test dataset using Rouge recall score with respect to the abstractive ground truth at 75 bytes and at 275 bytes. }
\end{table*}

Table \ref{tab:compare-baseline}  shows the performance comparison of our model with other baselines on the DailyMail dataset with respect to Rouge score at 75 bytes and 275 bytes of summary length.  
Our model performs consistently and significantly better than other models on 75 bytes, while on 275 bytes, the improvement margin is smaller.
One possible interpretation is that our model has high precision on top rank outputs, but the accuracy is lower for lower rank sentences. 
In addition, \cite{Cheng2016Neural} used additional supervised training to create sentence-level extractive labels to train their model, while our model uses an unsupervised greedy approximation instead.

\begin{table}[t!]
	\setlength{\abovecaptionskip}{0.cm}
	
	\setlength{\belowcaptionskip}{-0.cm}
	\begin{center}
		\setlength{\tabcolsep}{1.3mm}{
		\begin{tabular}{l|ccc }
			\hline CNN &  Rouge-1 &  Rouge-2 &  Rouge-L\\ 
			\hline
			Lead-3 & 29.1 & 11.1 & 25.9 \\
			Cheng \emph{et.al}'16  & 28.4 & 10.0 & 25.0 \\
			Hybrid MemNet&29.9& 11.3 & 26.4\\
			REFRESH &30.4 & 11.7 & 26.9\\
			ITS & \bf 30.8& \bf 12.6 & \bf 27.6 \\
			\hline
		\end{tabular}}
	\end{center}
	\caption{\label{cnn} Comparison with other baselines on \textbf{CNN} test dataset using full-length F1 variants of Rouge.  }
\end{table}

We also examined the performance of our model on CNN dataset as listed in Table \ref{cnn}. To compare with other models, we used full-length Rouge F1 metric as reported by \citet{Narayan2018Ranking}. Results demonstrate that our model has a consistently best performance on different datasets.

In Table \ref{tab:compare-duc}, we present the performance of ITS on the out of domain DUC dataset. 
Our model outperforms or matches other basic models including LReg and ILR as well as neural network baselines such as SummaRuNNer with respect to the ground truth at 75 bytes, which shows that our model can be adapted to different copora maintaining high accuracy. 
\begin{table}[t!]
	\setlength{\abovecaptionskip}{0.cm}
	
	\setlength{\belowcaptionskip}{-0.cm}
	\begin{center}
		\setlength{\tabcolsep}{1.3mm}{
		\begin{tabular}{l|ccc}
			\hline DUC2002 &  Rouge-1 &  Rouge-2 &  Rouge-L\\ 
			\hline
			Lead-3 & 43.6& 21.0 & 40.2 \\
			LReg & 43.8 & 20.7 & 40.3 \\
			ILP&45.4 & 21.3 & 42.8\\
			Cheng \emph{et.al}'16  & 47.4 & 23.0 & \bf 43.5 \\
			SummaRuNNer & 46.6 & 23.1 & 43.0 \\
			Hybrid MemNet&46.9& 23.0 & 43.1\\
			ITS & \bf 47.6& \bf 23.4 & \bf 43.5 \\
			\hline
		\end{tabular}}
	\end{center}
	\caption{\label{tab:compare-duc} Comparison with other baselines on \textbf{DUC2002} dataset using Rouge recall score with respect to the abstractive ground truth at 75 bytes.}
\end{table}


\begin{table}[t!]
	\begin{center}
		\setlength{\tabcolsep}{0.01mm}{
			\begin{tabular}{l|ccc}
				\hline Variations &  Rouge-1 &  Rouge-2 &  Rouge-L\\ 
				\hline
				ITS &  27.4&  11.9 &  16.1 \\
				\hline
				w/o selective reading & 27.1 & 11.6&15.4 \\
				w/o iteration & 26.9 & 11.6 & 15.8 \\
				w/o concatenation  & 27.2 & 11.7 & 15.9\\
				\hline
		\end{tabular}}
	\end{center}
	\caption{\label{three} Ablation study on DailyMail test dataset with respect to the abstractive ground truth at 75 bytes.}
\end{table}
In order to explore the impact of internal structure of ITS, we also conducted an ablation study 
in Table \ref{three}. The first variation is the same model without the selective reading module. The second one sets the iteration number to one, that is, a model without iteration process. The last variation is to apply MLP on the output from the last iteration instead of concatenating the hidden states of all decoders.
All other settings and parameters are the same.
Performances of these models are worse than that of ITS in all metrics, which demonstrates the preeminence of ITS. 
More importantly, by this controlled experiment, we can verify the contribution of different module in ITS.

\section{Further discussion}
\label{greetings}

\begin{figure}
	\centering
	\includegraphics[width=0.95\columnwidth]{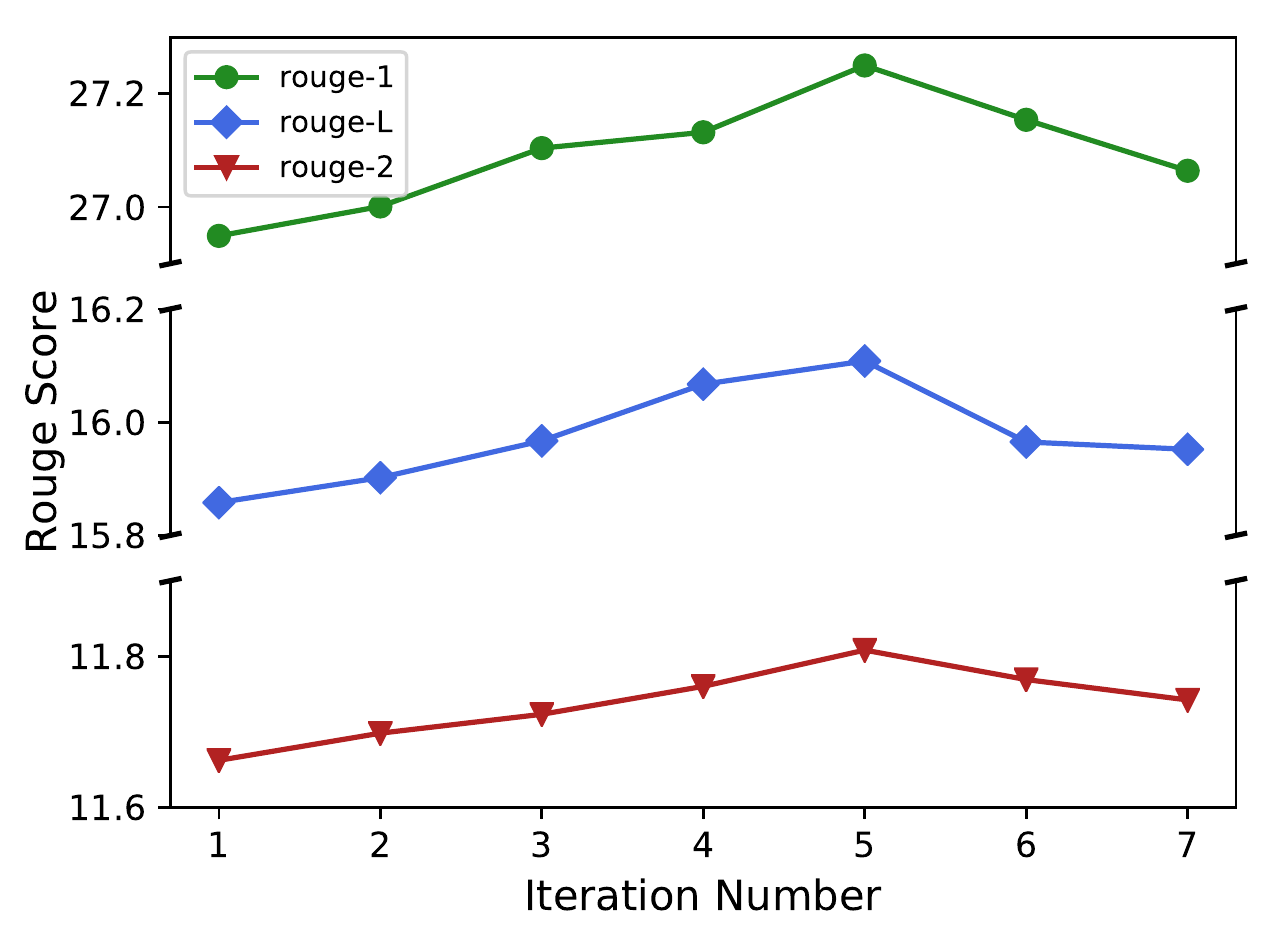}
	\caption{\label{fig:rouge-hop} Relationship between number of iteration and Rouge score on DailyMail test dataset with respect to the ground truth at 75 bytes.}
\end{figure}
\textbf{Analysis of iteration number:} We did a broad sweep of experiments to further investigate the influence of iteration process on the generated summary quality. 
First, we studied the influence of iteration number.
In order to make a fair comparison between models with different iteration number, we trained all models for same epochs without tuning. 
Fig.\ref{fig:rouge-hop} illustrates the relationship between iteration number and the Rouge score at 75 bytes of summary length on DailyMail test dataset. 
The result shows that the Rouge score increases with the number of iteration to begin with.
After reaching the upper limit it begins to drop.
Note that the result of training the model for only one epoch outperforms the state-of-the-art in \cite{Singh2017Hybrid}, which demonstrates that our selective reading module is effective.
The fact that continuing this process increase the performance confirms that the iteration idea behind our model is useful in practice. Based on above observation, we set the default iteration number to be 5.

\begin{figure} 
	\setlength{\abovecaptionskip}{0.cm}
	
	\setlength{\belowcaptionskip}{-0.cm}
	\setlength{\abovecaptionskip}{0.cm}
	
	\setlength{\belowcaptionskip}{-0.cm}
	\centering
	\subfigure[] { \label{fig:logits1}     
		\includegraphics[width=1.02\columnwidth]{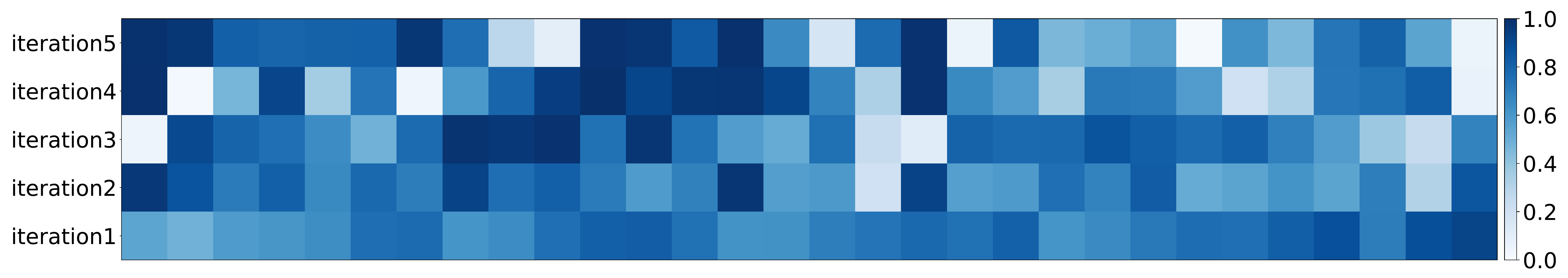}
	}       
	\subfigure[] { \label{fig:logits2}
		\includegraphics[width=1.02\columnwidth]{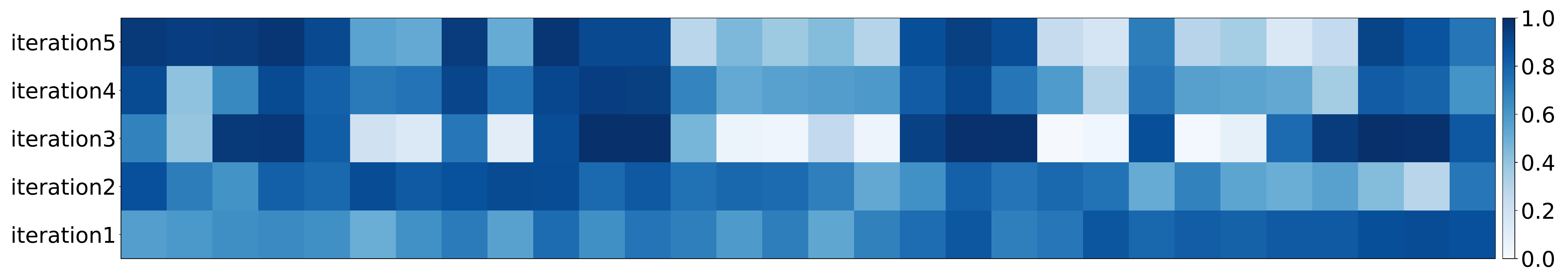}
	}  
	\caption{\label{fig:logits} The predicted extracting probabilities for each sentence calculated by the output of each iteration.} 
\end{figure}

\textbf{Analysis of polishing process:}
Next, to fully investigate how the iterative process influences the extracting results, we draw heatmaps of the extracting probabilities for each decoder at each iteration. We pick two representative cases in Fig.\ref{fig:logits}, where the $x$-axis represents the sentence index and $y$-axis is the iteration number, $x$-axis labels are omitted.
The darker the color is, the higher the extracting probability is. 
In Fig.\ref{fig:logits}(a), it can be seen that when the iteration begins, most sentences have similar probabilities.
As we increase the number of iteration, some probabilities begin to fall and others saturate. This means that the model already has preferred sentences to select.
Another interesting feature we found is that there is a transitivity between iterations as shown in Fig.\ref{fig:logits}(b). 
To be specific, the sentences which are not preferred by iteration 3 remain low probabilities in the next two iterations, while sentences with relatively high scores are still preferred by iteration 4 and 5.

\textbf{Human Evaluation:}
We gave human evaluators three system-generated summaries, generated by Lead-3, Hybrid MemNet, ITS, as well as the human-written gold standard summary, and asked them to rank these summaries based on summary informativeness and coherence. Table \ref{tab:model}  shows the percentages of summaries of different models under each rank scored by human experts.  
It is not surprising that gold standard has the most summaries of the highest quality. 
Our model has the most summaries under 2nd rank, thus can be considered 2nd best, following are Hybrid MemNet and Lead-3, as they are ranked mostly 3rd and 4th.
By case study, we found that a number of summaries generated by Hybrid MemNet have two sentences the same as ITS out of three, however, the third distinct sentence from our model always leads to a better evaluation result considering overall informativeness and coherence.
Readers can refer to the appendix to see our case study.

\begin{table}[t!]
	\setlength{\abovecaptionskip}{0.cm}
	
	\setlength{\belowcaptionskip}{-0.cm}
	\begin{center}
		\begin{tabular}{l|cccc}
			\hline Models &  1st &  2nd &  3rd & 4th\\ 
			\hline
			Lead-3 & 0.12 &0.11 & 0.25 & \bf 0.52\\
			Hybrid MemNet& 0.24 &0.25 &\bf 0.28 & 0.23\\
			ITS & 0.31 &\bf 0.34 & 0.23 & 0.12\\
			Gold & \bf0.33 &0.30 & 0.24 & 0.13\\
			\hline
		\end{tabular}
	\end{center}
	\caption{\label{tab:model} System ranking comparison with other baselines on DailyMail corpus. Rank 1 is the best and Rank 4 is the worst. Each score represents the percentage of the summary under this rank.}
\end{table}

\section{Conclusion}
In this work, we introduce ITS, an iteration based extractive summarization model, inspired by the observation that it is often necessary for a human to read the article multiple times to fully understand and summarize it. 
Experimental results on CNN/DailyMail and DUC corpora demonstrate the effectiveness of our model.

\section*{Acknowledgments}

We would like to thank the anonymous reviewers for their constructive comments. We would also like to thank Jin-ge Yao and Zhengyuan Ma for their valuable advice on this project. This work was supported by the National Key Research and Development Program of China (No. 2017YFC0804001), the National Science Foundation of China (NSFC No. 61876196, No. 61672058). Rui Yan was sponsored by CCF-Tencent Open Research Fund and Microsoft Research Asia (MSRA) Collaborative Research Program.

\bibliography{ms}
\bibliographystyle{acl_natbib}

\end{document}